\documentclass[journal]{IEEEtran}
\RequirePackage{subfigure}

\usepackage{hyperref} 
\usepackage{times}
\usepackage{epsfig}
\usepackage{graphicx}
\usepackage{amsmath}
\usepackage{amssymb}
\graphicspath{ {images/} }
\usepackage{colortbl}
\usepackage[table]{xcolor}

\usepackage{array,multirow}
\usepackage{caption}

\definecolor{Gray}{gray}{0.8}

\begin{document}
%
\title{Identifying Object States in Cooking-Related Images}

\author{Ahmad~Babaeian~Jelodar,~Md~Sirajus~Salekin,~and~Yu~Sun
\thanks{A. B. Jelodar, D. Paulius, and Y. Sun are with the Department
of Computer Science and Engineering, University of South Florida, Florida
FL, 33620 e-mail: (ajelodar@mail.usf.edu).}
}

\maketitle

\begin{abstract}
Understanding object states is as important as object recognition for robotic task planning and manipulation. 
To our knowledge, this paper explicitly introduces and addresses the state identification problem in cooking related images for the first time. 
In this paper, objects and ingredients in cooking videos are explored and the most frequent objects are analyzed. 
Eleven states from the most frequent cooking objects are examined and a dataset of images containing those objects and their states is created. 
As a solution to the state identification problem, a Resnet based deep model is proposed.
The model is initialized with Imagenet weights and trained on the dataset of eleven classes.
The trained state identification model is evaluated on a subset of the Imagenet dataset and state labels are provided using a combination of the model with manual checking.
Moreover, an individual model is fine-tuned for each object in the dataset using the weights from the initially trained model and object-specific images, where significant improvement is demonstrated.

\end{abstract}


\IEEEpeerreviewmaketitle

\section{Introduction}
Image understanding for object recognition and scene understanding have been very active topics in the last few years \cite{Resnet,Review_ActionRecognition1,Review_ImageCaption1}. On the other hand, identifying object states has not captured much attention in computer vision and robotics research. 
In this study, we defined objects \textit{states} as characters into which the object could be transformed as a consequence of a human or robot activity. 
A state can be observed and described as a form, texture, or color. For example, a tomato can have many states, such as sliced, diced, and whole. A whole tomato can be sliced and then diced in a sequence of cooking activity such as slicing and dicing. Assuming a robot chef wants to make a salad using a tomato, if it is provided with a whole tomato, it would need to wash it, slice it, and then dice it. If it is provided with a sliced tomato to begin with, it would need only to dice it. The intelligent robot chef would need to plan its motion differently based on the state of the provided tomato. Therefore, it would be necessary to not only recognize the object as a tomato, but also to identify the state the tomato is in. This is important for both fine-grained human activity understanding and robot task planning and manipulation control. 

Robots also need to perform different manipulations or grasps to achieve different states of a planned task \cite{lin2016task,lin2015task,lin2015grasp,lin2015robot}. Different states of an object or transiting an object from one state to another requires different types of grasping; for example, a whole carrot is grasped differently than a sliced or grated carrot \cite{sun2016robotic,lin2012learning}, or holding a whole carrot for slicing, holding a half carrot for grating, or holding a julienne-cut carrot for dicing each need unique types of grasping. Receiving on-line feedback from the environment would give the robot the sufficient knowledge required to decide on the unique type of grasp it would choose for its manipulation of the environment.

In this paper, we present our exploration on identifying object states in cooking-related images. First, we selected 17 of the most commonly-used cooking objects from more than 250 online cooking videos of two of the well known cooking datasets \cite{FOON,MaxPlankIICooking} and identified their states in the videos, resulting in 11 different states for all 17 objects. Subsequently, we created our own state identification dataset of 9309 images of these objects and their state labels. 
Using the dataset, we built and trained a Resnet-based deep neural network model starting from pre-trained weights. 
We evaluated our approach with the images in the ImageNet \cite{ImageNet} and then assigned the images their state labels using a combination of our state identification model and manual labeling. 

Our work has three main contributions:
\begin{itemize}
  \item We define the state identification problem in cooking for fine-grained activity understanding and robot manipulation and provide a labeled dataset for the state identification problem.
  \item We designed a Resnet-based deep architecture and a transfer learning approach that used the pre-trained weights and a small number of labeled images for each state of all objects and then a much lower number of labeled images of each state of every object. 
  \item We provide state labels for the images in the Imagenet dataset containing those 17 objects. \end{itemize}

The rest of the paper is organized as follows. Section \ref{pre_work} discusses related work, and Section \ref{Section_challenge} introduces the state identification challenge and describes how data are collected into a state dataset. Section \ref{section_approach} introduces the algorithm proposed for state identification, and Section \ref{section_experiments} discusses experiments and results. Further discussion is provided in Section \ref{section_discussion}, and Section \ref{section_Conclusion}.

\section{Related Work} \label{pre_work}
\label{pre_work}

Some research has been done in the area of state recognition such as 
\cite{added4_objectstates} and \cite{added1_multi_task_cnn}. Some of the work approach the problem jointly with action classification \cite{added3_StatesAndTransformations} and some perform state identification implicitly \cite{added2_densecap}.
To the best of our knowledge, no specific work has been done in the area of image state identification on cooking images. 
In this section, we discuss work in the area of image classification, image captioning, and understanding that are relevant or similar to state identification or motivated us for this research.
Currently, image classification has shifted towards convolutional neural networks. 
Krizhevsky et al, introduced the first evolutionary deep model for image classification \cite{Alexnet}. 
Thereafter, other deep models such as VGG \cite{VGG}, Googlenet \cite{Googlenet}, and Resnet \cite{Resnet} were introduced gradually as deeper and more advanced networks for image classification. 
Improvements with the combination of these networks have also has been introduced in \cite{improvements}. 
These works all focus on image object classification and do not consider states of objects in an image. 

In \cite{Review_ActionRecognition1}, the authors show the importance of using object parts in recognizing an action from an image, thereby modeling human actions based on parts and attributes in an image. 
This work is an obvious proof of how object parts and states can help recognize an object or understand an image. 
Researches such as \cite{Review_ImageCaption1} and \cite{Review_ImageCaption2} provide captions for images or videos. In \cite{Review_ImageCaption1} Yao et al. use attributes and their interactions with deep networks to provide captions. 
Other work such as \cite{Review_ImageCaption3}, and \cite{Review_ImageCaption4} perform multi-label classification on a single image using RNN- and CNN-based deep architectures. Although these papers provide various labels for an image, they do not consider states of objects as another label for the image. These papers have one thing in common -- they analyze an image to understand it. The state identification problem, also motivated by this aspect, contributes to the understanding of images.

Besides work in image classification and recognition, some work has been conducted in the area of cooking images and videos. 
Food recognition systems for dietary analysis \cite{Review_Cooking1,Review_Cooking3}, fruit recognition \cite{Review_Cooking2}, and ingredient recognition for recipe retrieval \cite{Review_cooking8} are instances of work, which focus on recognizing or detecting the ingredients in an image. 
Some papers, such as \cite{Review_Cooking5,Review_cooking7}, perform food recognition on video to understand the whole video and associate it with a recipe or action. Other papers, such as \cite{Review_cooking10,Review_cooking6}, focus on activity recognition from cooking videos. 
These works contribute to understanding cooking images and videos, but none explicitly focus on states. To our knowledge, we are the first to address this problem in cooking images or videos.

\section{The state identification Challenge}
\label{Section_challenge}

In this section, the state identification challenge and the dataset collected for the challenge are introduced, and the data collection procedure, dataset statistics, and details of the dataset are discussed. 

\subsection{The Challenge} \label{subsection_challenge}

In our daily lives, we perform tasks by paying attention to objects and their states and how they interact with each other. Like humans, one of the main tasks of an intelligent robot is to properly manipulate the environment. For a smart robotic system to perfectly manipulate the environment, it needs to acquire accurate knowledge of the environment, objects, and their affordances and status. 

An object can contain various shapes and states, therefore introducing various ways of manipulation. For example, when making an omelette, we need to dice peppers and onions. 
To dice a pepper, we need to grab the whole pepper, place it on a cutting board, use a knife to cut it in half, and, finally, cut it into julienne cuts and dice the cuts into small pieces. 
We can observe that the simple action of dicing a pepper requires knowledge of four different states for us or an intelligent robot. 
As humans, we get constant feedback from the objects (pepper). 
A robot needs to also gain feedback from the state of the pepper to decide how to continue the cutting. 
Knowing the current status of an object helps the robot with how it approaches the manipulation of the object. 
In this example, for simplicity, we classified the states of an object into 4 different states, but in a real-world environment, the actual states of an object are continuously changing. 

This example demonstrates the need for classification of an object (pepper) into a diverse set of states (whole, half, julienne, diced). Thus, we introduce the \textbf{\textit{state identification challenge}} for the first time in this paper. We define a state of an object (such as a tomato) as the various physical shapes (diced, paste, juice, or whole) into which the object can be transformed as a consequence of human or robot activity. We propose and anticipate that by solving the state identification challenge, we can step towards accurately understanding and executing robot manipulation tasks such as grasping.

\begin{figure*} [!h]
   \centering
   \includegraphics[width=16cm]{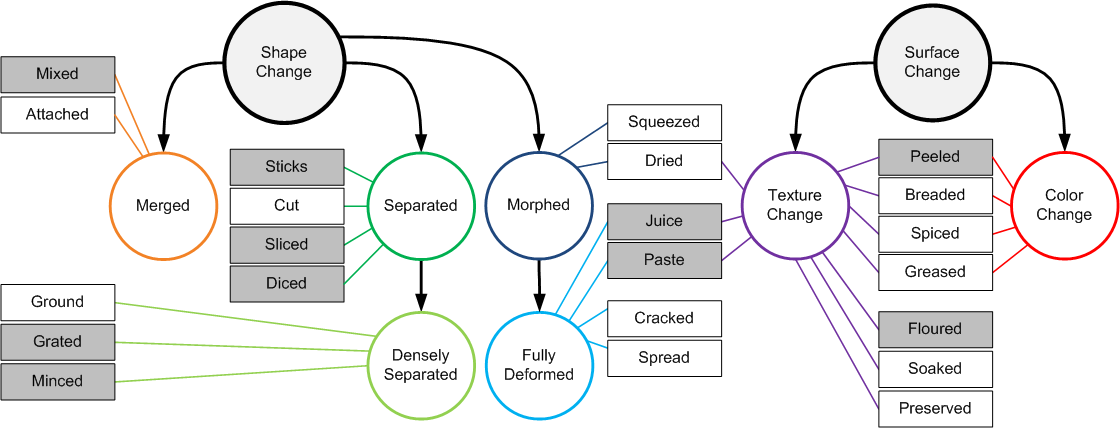}
   \caption{state identification problem definition.}
\label{fig:state_idea}
\end{figure*}

One of the main problems and applications of robotic systems is the cooking scope. 
In this study, we designed the state identification challenge for cooking objects. 
We analyzed cooking objects and their states by looking into the statistics extracted from the knowledge representation introduced in \cite{FOON}. We discerned the most frequent objects in this knowledge representation and explored their states. 
State analysis shows that there are two major states for cooking objects-shape change and surface change. 
Hierarchical exploration shows that there are three main states under shape change, namely separated, morphed, and merged, each of which can further be represented by finer states. 
Also, surface change can be divided into two states, color and texture change, which, in turn, have finer representations themselves. 
Figure \ref{fig:state_idea} depicts the hierarchical representation of explored states. In some cases, an object can have a combination of multiple states simultaneously.
In this study, for simplicity, we assumed that an object can have only a single state at a time (in one snapshot).

A total of 22 fine states is shown in Figure \ref{fig:state_idea}. These states represent the whole state space that all important objects from \cite{FOON} can span. 
We selected only 10 states (shown in gray in the figure) that are representative of the whole state space for our problem. 
The main reason to this state reduction is the lack of training image samples for the eliminated states. 
We can represent the state space in any other scope with a similar graph and further analyze the problem in that scope. 
In this study, we focused only on the cooking state space. 

\subsection{The States Identification Dataset and Statistics}

Dataset images were crawled through the Google search engine using a keyword combination of each object and state (such as ``tomato'' and ``sliced''). 
The links to the crawled images were exported to a file, downloaded, and reviewed to remove unrelated images (such as cartoon images). 
Using the Vatic annotation tool, \cite{VATIC_tool}, images were published through a local server and dispersed to multiple workers for manual labeling. The labels were further reviewed and were gathered into a dataset of states. 
A small fraction of the dataset was labeled through the labelbox tool \cite{Labelbox}.

\begin{figure} 
   \centering
   \includegraphics[width=8cm]{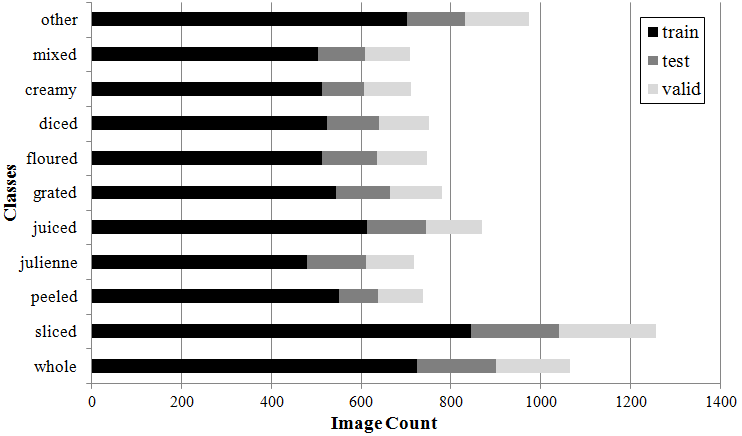}
   \caption{Dataset class statistics.}
\label{fig:dataset_stats}
\end{figure}

The collected dataset consisted of 17 main cooking objects, including tomato, onion, garlic, green pepper, potato, carrot, strawberry, egg, mushroom, bread, beef/pork, chicken/turkey, cheese, butter, dough and milk and 11 classes of states (whole, peeled, floured, sliced, diced, grated, julienne, juice, creamy, mixed, other). 
The total number of images in the dataset is 9309 -- 6498(70\%) training, 1413(15\%) testing, and 1398(15\%) validation images. The statistics of each class in the dataset is depicted in Figure \ref{fig:dataset_stats}. The classes ``whole'' and ``sliced'' contain more than 1000 images, and the other classes contain approximately 700 to 1000 images.

\subsection{The Dataset details}

\begin{figure*} [!h]
\centering
\includegraphics[width=17cm]{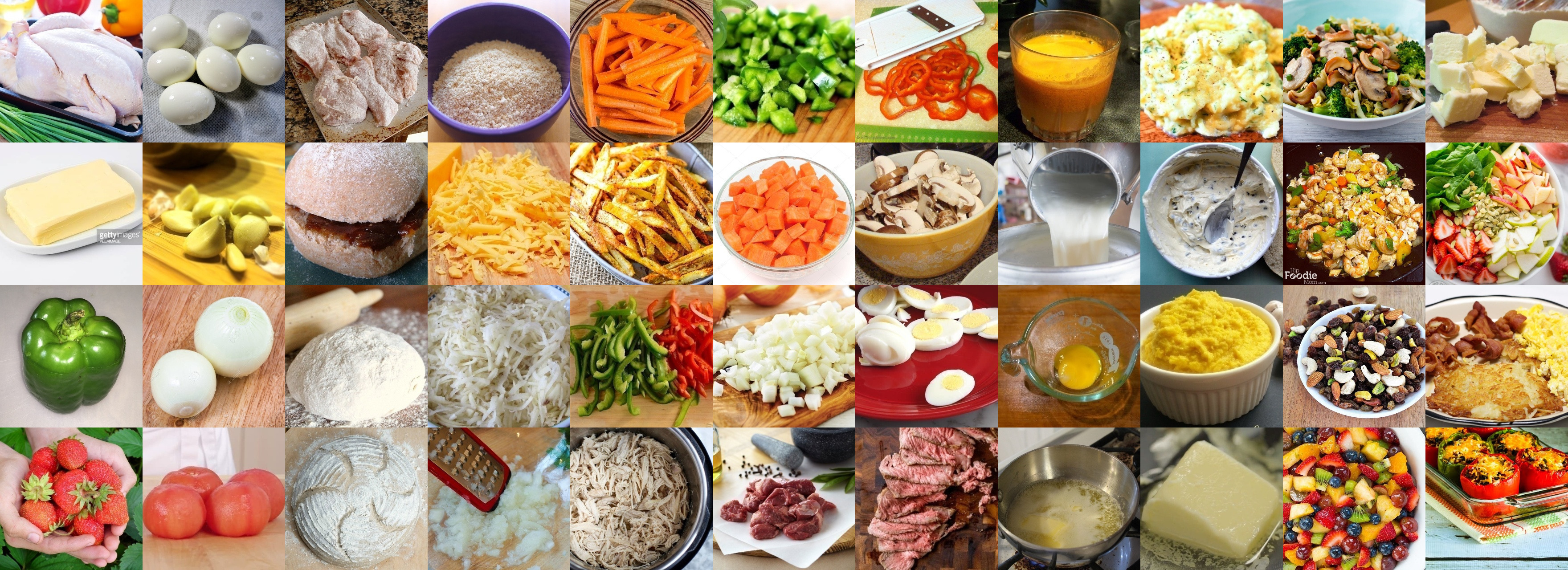}
\caption{Example images of 11 classes of the Dataset.}

\label{fig:dataset_images}
\end{figure*}

In this section, we give a concise definition of each of the 11 states to clearly state what each state represents. 
The \textit{whole} state contains objects in their original format and shape, such as a whole pepper or a whole chicken, as shown in the first column of Figure \ref{fig:dataset_images}. 
The \textit{peeled} state contains objects that are peeled but not cut, sliced, or morphed, such as a whole peeled egg, onion, garlic, or tomato, (shown as in column 2 of Figure \ref{fig:dataset_images}).
The \textit{floured} state as depicted in column 3 of Figure \ref{fig:dataset_images} contains objects that are floured. 
The \textit{grated} state comprises of objects that are densely separated, such as bread crumbs, minced garlic, or grated egg (column 4 of Figure \ref{fig:dataset_images}).
The \textit{julienne} state includes objects such as carrot sticks, French fries, julienne pepper, or shredded meat (column 5 of Figure \ref{fig:dataset_images}).

The \textit{diced} state contains diced or chopped objects, such as diced onion, tomato, and strawberry, chopped meat, and butter, and cheese cubes. Examples of this state are shown in column 6 of Figure \ref{fig:dataset_images}. 
The \textit{sliced} class contains objects that are thinly-sliced such as sliced carrot, pepper, onion, tomato, meat or chicken slices, toast, and butter and cheese slices (Figure \ref{fig:dataset_images}, column 7). 
Objects that are cut in other ways (such as cut in half or diced) are not considered as sliced.
The \textit{juiced} class contains objects such as milk, melted butter, and tomato juice, (column 8 of Figure \ref{fig:dataset_images}). 
The \textit{creamy} state contains objects that are creamy, such as cream, creamy butter or cheese, garlic or tomato paste, and mashed potato (Figure \ref{fig:dataset_images}, column 9). 
The \textit{mixed} class contains a scramble of multiple objects such as salads (Figure \ref{fig:dataset_images}, column 10).  
A final class called the \textit{other} class is created that includes any state not listed in the previous states. 
A potato cut in half, squeezed lemon, images with multiple states, and an unmixed salad are assumed to be in this class (Figure \ref{fig:dataset_images}, column 11). 
Note that each object contains only a subset of the 11 states.

\section{The Methodology} \label{section_approach}

\begin{figure*}
\centering
\includegraphics[width=0.9\textwidth]{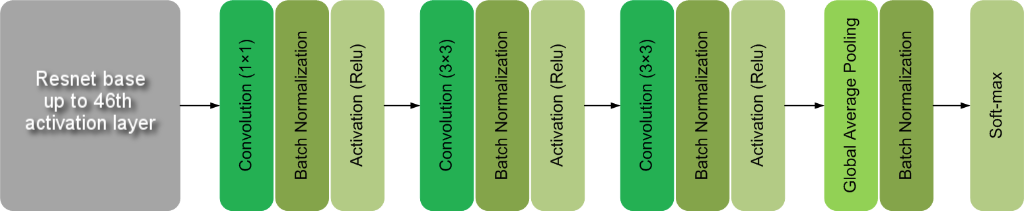}

\caption{Proposed Network structure.}

\label{fig:network_structure}
\end{figure*}

The state identification problem is an image classification problem. 
Like other recent image classification problems in the last few years, we propose to solve the problem with a deep structure. Our model uses the Resnet base model up to the 46th activation layer \cite{Resnet} as its basis. 
We added a layer of 1x1 convolution, two layers of convolution, and a layer of global averaging before the 11 class soft-max layer. We used batch normalization in each layer for normalization and regularization purposes in the network. 
The structure of the network is depicted in Figure \ref{fig:network_structure}. The 1x1 convolution was added to make the feature map set shallower. The convolutions were added to capture new spatial features for the specific state identification problem. 

\subsection{Transfer Learning} \label{subsection_transfer}

The network included more than 19 million parameters; therefore, the Resnet base, pre-trained on Imagenet, was used for training the state identification model. The pre-trained weights of the model initially were frozen in the first step and, further in the training procedure, the whole network (with all the parameters and 11 classes) was fine-tuned. More details about the transfer learning procedure are given in subsection \ref{section_experiment1}.

\subsection{Object Specific Fine-tuning} \label{subsection_tuning}

State identification has a strong correlation with the type of object. For instance, butter cannot be found as grated, julienne, or peeled and rarely can be seen as diced. On the other hand, cheese can be grated and lemon can be zested. We took this knowledge into consideration by fine-tuning 17 individual models for each object in the dataset. Each object has a different number of states; for example, garlic has 5 states including whole, paste, minced, peeled, and sliced, but carrot can have 7 states including julienne, diced, and juice. Therefore, the last layer of Figure \ref{fig:network_structure} was removed and a new soft-max layer was added to the model. 
The number of units in the new soft-max layer is equivalent to the number of states that an individual object could have. 
The number of images for each object in the dataset was limited; therefore, we do transfer learning; we initially trained the network on the whole dataset and fine-tune the modified network on a small number of object-specific images. 
These fine-tuned networks can be used in a hierarchical manner after a network is used for object classification.
This will increase the accuracy for state classification.

\section{Experiments and Results}  \label{section_experiments} 
We designed three experiments. In the the first, we trained and tested the deep architecture on the whole dataset. 
In the second, we showed that fine-tuning the model individually for each object improves the accuracy for state identification.
In the third, we tested our model on a sample of Imagenet images and provided state labels for a subset of their dataset \cite{ImageNet}. 

\subsection{State Identification} \label{section_experiment1}

We performed two sets of experiments; state identification without and with prior object based fine-tuning. 
For state identification without individual object based fine-tuning, the model was trained and then evaluated on an unseen test set. 
We used the Adam optimizer with a learning rate of 0.001, beta1 of 0.9, and beta2 of 0.999, froze the Resnet base of the model, and trained only the layers we added to the network for 100 epochs. 
We then fine-tuned all layers of the model (including the Resnet base) for 250 epochs with a learning rate of 0.000005. On-line data augmentation, l2 regularization, and batch normalization was performed to reduce over-fitting. 
The average class accuracy calculated for the trained and tested sets are 81.4\% and 80.4\%, respectively (Table \ref{table_model_acc}).
We trained 2 other models using a Resnet base and similar architectures. 
Using the validation set, a weighted voting was performed between these 3 models, and the best combination of weights was used for the final model. 
As shown in Table \ref{table_model_acc}, after voting the state recognition accuracy rose to 82\%. 

\begin{table} [h]
\centering
\caption{Classification accuracy on the state dataset and Imagenet subset.}
\label{table_model_acc}
\scalebox{1.0}{
 \begin{tabular}{|c|c|c|c||c|c|}
 \hline
\multirow{2}{*}{} & \multirow{2}{*}{Model} & \multicolumn{2}{c||}{State Dataset} & \multicolumn{2}{c|}{Imagenet Subset} \\ 
 \cline{3-6}
  &  & Top 1 & Top 2 & Top 1 & Top 2 \\ [0pt]
 \hline
 1 & Resnet-based Model & 80.4\% & 91.5\% & 78.5\% & 89.6\% \\ [0pt]
  \hline
 2 & Voting & 82\% & 92\% & \multicolumn{2}{c|}{-} \\ [0pt]
  \hline
\end{tabular}
}
\end{table}

For fine-tuning the model for each individual object, we perform a 4 stage training.
In stage 1 and 2 all layers but the last are frozen.
In stage 3 our additional layers are unfrozen and in stage 4 the whole model is unfrozen. 
Learning rates for stages 1 to 4 are 0.01, 0.001, 0.00001, and 0.000005 respectively. Epochs for stages 1 to 4 are 40, 80, 120, and 160 respectively. 
The First 5 columns of Table \ref{table_fine_tune} shows the classification accuracies of the fine-tuned model for each object.
The object dough was removed from this set of experiments.

\begin{table} [h]
\centering
\caption{Classification accuracy of the individual fine-tuning}
\label{table_fine_tune}
\scalebox{1.1}{
 \begin{tabular}{|c|c|c|c|c|}
 \hline
  Object & Top 1 & Voting & States & Test Set \\ [0pt]
  \hline
 mushroom & 95.6\% & 97.8\% & 3 & 45 \\ [0pt]
 \hline
 onion & 80.2\% & 85\% & 7 & 86 \\ [0pt]
 \hline
 strawberry & 92.6\% & 92\% & 4 & 68  \\ [0pt]
 \hline
 bread & 78.9\% & 78.9\% & 6 & 123 \\ [0pt]
 \hline
 butter & 69.7\% & 72.7\% & 5 & 66 \\ [0pt]
 \hline
 carrot & 78.5\% & 84.9\% & 8 & 135  \\ [0pt]
 \hline
 egg & 90.6\% & 89.2\% & 5 & 85  \\ [0pt]
 \hline
 garlic & 86.7\% & 85.3\% & 5 & 75  \\ [0pt]
 \hline
 lemon & 90.7\% & 94.9\% & 6 & 108 \\ [0pt]
 \hline
 milk & 100\% & 100\% & 2 & 40 \\ [0pt]
 \hline
 pepper & 96.1\% & 97.5\% & 5 & 76 \\ [0pt]
 \hline
 potato & 84\% & 88.3\% & 8 & 106 \\ [0pt]
 \hline
 tomato & 88.5\% & 91.1\% & 7 & 113 \\ [0pt]
 \hline
 cheese & 82.7\% & 78.7\% & 4 & 75  \\ [0pt]
  \hline
  \rowcolor{Gray}
 beef/pork & 86.7\% & 86.7\% & 5 & 60 \\ [0pt]
 \hline
 \rowcolor{Gray}
 chicken & 88.8\% & 89.7\% & 6 & 116 \\ [0pt]
 \hline
  average & 86.9\% & 88.3\% & 5.4  & 86.1 \\ [0pt]
  \hline
\end{tabular}
}
\end{table}

\subsection{ImageNet Test} \label{section_experiment2} 

In this experiment, we contributed to the Imagenet dataset by providing state labels for them.
For each object category in our dataset, excluding beef and chicken, 50 images were randomly selected from the Imagenet dataset. Beef and chicken were excluded because Imagenet does not contain cooking-related images for these two categories. 
The Salad synset was included for the experiments because it is considered a frequent image in cooking videos, thus leading to a total of 16 object categories and 800 images. The images were labeled with the 11 classes in our dataset. 

Moreover, the trained model was run on the Imagenet subset and the average state identification accuracy on the Imagenet subset was reported as 78.5\%. Individual accuracies for the top 1, 2 and 3 are shown in the last three columns of Table \ref{table_imagenet}.
In addition to the evaluation, we ran our model on all images associated to the 16 object categories and gave their labels. Then we  manually checked the labels through an interface and keep the correct labels and discard the others. The state labels of the images and the dataset will be released on our website for download after the double-blind peer review process.



\begin{table} [h]
\centering
\caption{Classification accuracy of the Imagenet subset}
\label{table_imagenet}
\scalebox{1.1}{
 \begin{tabular}{|c|c|c|c|}
 \hline
Object & Top 1 & Top 2 & Top 3 \\ [0pt]
  \hline
 mushroom & 74\% & 84\% & 96\% \\ [0pt]
 \hline
 onion & 86\% & 94\% & 96\% \\ [0pt]
 \hline
  strawberry & 74\% & 86\% & 90\% \\ [0pt]
 \hline
 bread  & 80\% & 96\% & 98\% \\ [0pt]
 \hline
 butter & 70\% & 88\% & 94\% \\ [0pt]
 \hline
 carrot & 96\% & 100\% & 100\% \\ [0pt]
 \hline
  egg & 62\% & 78\% & 86\% \\ [0pt]
 \hline
 garlic & 84\% & 90\% & 92\% \\ [0pt]
 \hline
 orange & 90\% & 96\% & 98\% \\ [0pt]
 \hline
  milk & 72\% & 80\% & 88\% \\ [0pt]
 \hline
  pepper & 78\% & 88\% & 94\% \\ [0pt]
 \hline
 potato & 72\% & 94\% & 96\% \\ [0pt]
 \hline
 tomato & 74\% & 82\% & 92\% \\ [0pt]
 \hline
 cheese & 84\% & 90\% & 98\% \\ [0pt]
  \hline
  \rowcolor{Gray}
 salad & 90\% & 98\% & 98\% \\ [0pt]
 \hline
 \rowcolor{Gray}
 dough & 70\% & 90\% & 96\% \\ [0pt]
 \hline
  average & 78.5\% & 89.6\% & 94.5\% \\ [0pt]
  \hline
\end{tabular}
}
\end{table}

\section{Discussion}\label{section_discussion} 

\subsection{State Analysis} \label{section_DiscussionA}

Classification accuracy of all classes apart from the \textit{other} class is at least 70\% for the experiment on our test set.
The other class includes various kinds of images, such as images of meals, sandwiches, images including a combination of various states (classes), and etc. 
The variety in the other class is the reason of such a low accuracy. 
We anticipate performing a joint detection and recognition procedure of states would improve the accuracy of classification in all classes.

\begin{figure} [!h]
\centering     
\subfigure[Crumbs as floured]{\label{fig:a}\includegraphics[width=30mm]{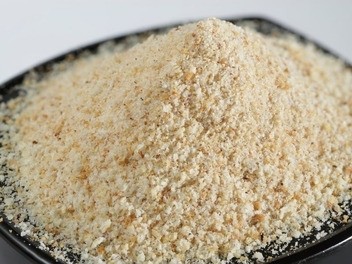}}
\hspace{2mm}
\subfigure[Sliced as creamy]{\label{fig:b}\includegraphics[width=30mm]{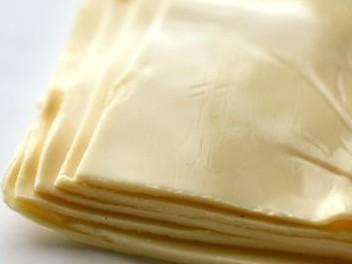}}
\hspace{2mm}
\subfigure[Melted as juice]{\label{fig:c}\includegraphics[width=30mm]{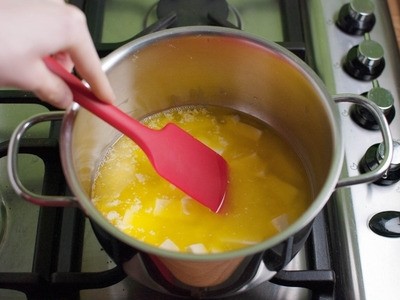}}
\hspace{2mm}
\subfigure[Grated as Julienne]{\label{fig:d}\includegraphics[width=30mm]{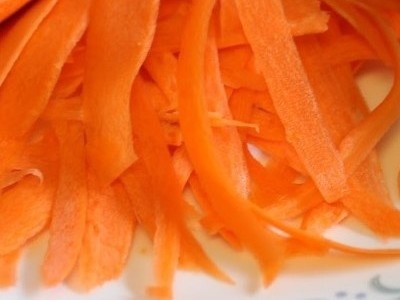}}

\subfigure[mis-labeled]{\label{fig:e}\includegraphics[width=30mm]{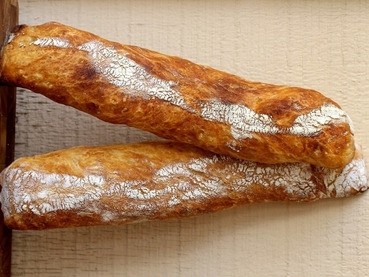}}
\hspace{2mm}
\subfigure[mis-labeled]{\label{fig:f}\includegraphics[width=30mm]{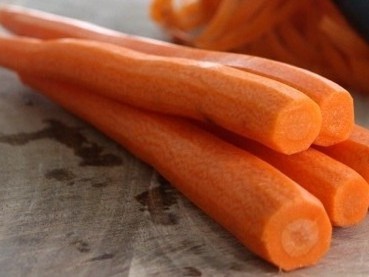}}
\hspace{2mm}
\subfigure[Multi-state]{\label{fig:g}\includegraphics[width=30mm]{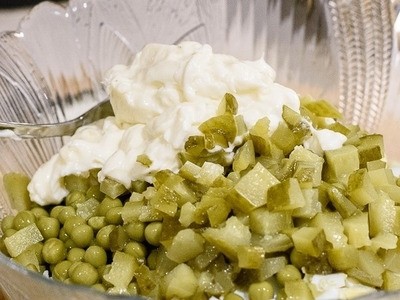}}
\hspace{2mm}
\subfigure[Multi-state Carrot]{\label{fig:h}\includegraphics[width=30mm]{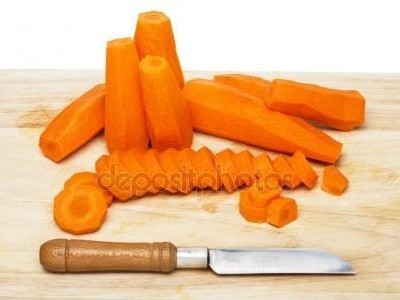}}
\caption{Samples of mis-predicted (a, b), ambiguous (c, d), mis-labeled (e, f) and multi-state (g, h) images. (e) is floured and (f) is peeled but they are mis-labeled as whole in the dataset. }
\label{fig:ambiguous_our}
\end{figure}

A majority of the mistakes made by the model were on account of ambiguous and multi-state images as depicted in Figure \ref{fig:ambiguous_our}.
This also suggests that detection of all states inside an image rather than looking at the entire image as a whole may improve the state identification accuracy. 
Moreover, tracking the state of an object and assigning values for representing the quality of an object being in a specific state may improve state identification results.

\subsection{Imagenet Analysis} \label{section_DiscussionB} 

\begin{figure} [!h]
\centering     
\subfigure[Pepper as crop]{\label{fig:a}\includegraphics[width=30mm]{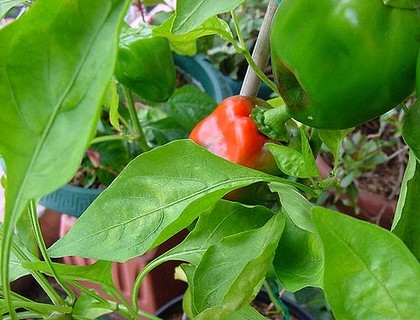}}
\hspace{2mm}
\subfigure[Strawberry as crop]{\label{fig:b}\includegraphics[width=30mm]{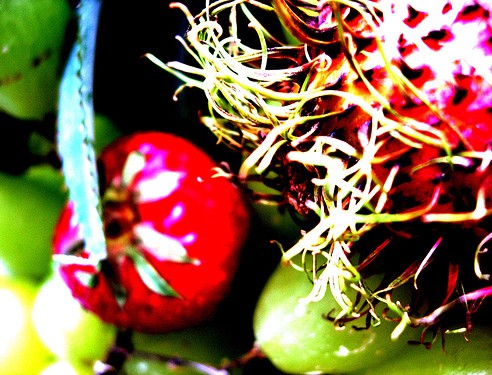}}
\hspace{2mm}
\subfigure[Peeled pepper]{\label{fig:c}\includegraphics[width=30mm]{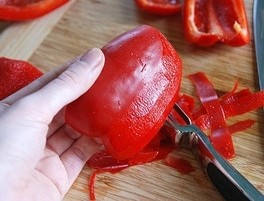}}
\hspace{2mm}
\subfigure[Yolk and grated]{\label{fig:d}
\includegraphics[width=30mm]{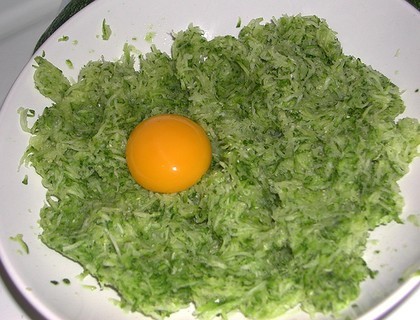}}
\hspace{2mm}
\subfigure[Butter and jelly]{\label{fig:d}
\includegraphics[width=30mm]{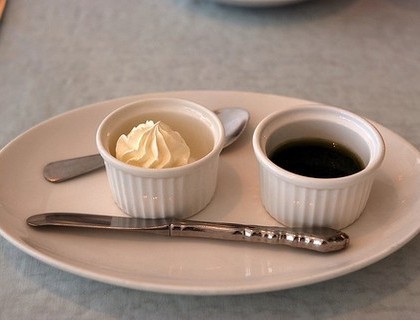}}
\hspace{2mm}
\subfigure[Yolk and cracked]{\label{fig:d}
\includegraphics[width=30mm]{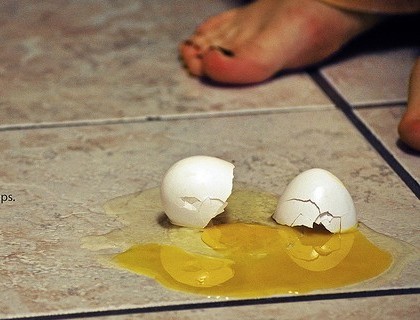}}
\hspace{2mm}
\subfigure[Whole and yolk]{\label{fig:d}
\includegraphics[width=30mm]{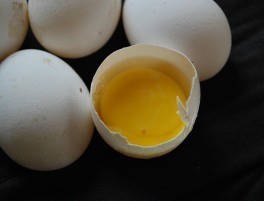}}
\hspace{2mm}
\subfigure[Potato as dish]{\label{fig:d}
\includegraphics[width=30mm]{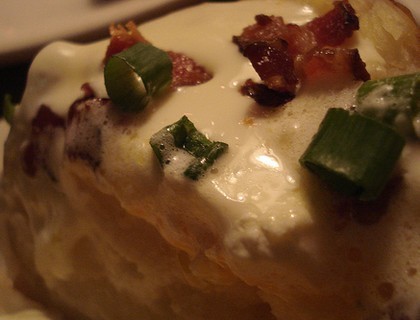}}
\caption{Samples of multi-state or ambiguous images in Imagenet.}
\label{fig:ambiguous_imagenet}
\end{figure}

Unexpectedly, the average state identification accuracy on the Imagenet subset, despite having more ambiguous images, was only slightly lower than the classification accuracy on our dataset.
Figure \ref{fig:ambiguous_imagenet} shows examples of ambiguous images from the Imagenet subset.
These images either contain multiple states such as Figure \ref{fig:ambiguous_imagenet}.d. or 
are out of our dataset scope such as Figure \ref{fig:ambiguous_imagenet}.a.
Interestingly, the half peeled pepper, in Figure \ref{fig:ambiguous_imagenet}.c., was predicted as peeled although no image of a peeled pepper is included in our dataset. 
Figure \ref{fig:ambiguous_imagenet}.b. was counted as a wrong prediction, although the model's first three predictions for this image is whole, julienne and other. 
This example shows that the model is able to capture sufficient features, but does not have the tool to identify multiple states in an image simultaneously.

\section{Conclusion and Future Work} \label{section_Conclusion} 

In this paper, the state identification challenge for cooking images is introduced for the first time, and a solution to it is provided using a deep convolutional approach. 
A state of an object is defined as the form an object could be transformed into, and the state identification challenge is defined as the problem of classifying an image of an object into its relative state. 
A useful dataset of cooking ingredients was gathered for the challenge. Using a proposed deep model based on the Resnet \cite{Resnet} architecture, a promising level of accuracy was reached for state identification. 
We further tested our model on Imagenet images and semi-automatically provided state labels for images in Imagenet that are related to cooking ingredients. 
We showed that fine-tuning the model for each known object improves the average accuracy significantly. In future work, we will explore detection of all states inside an image, tracking the states in a video and providing continuous state labels for objects in a video.

\ifCLASSOPTIONcaptionsoff
  \newpage
\fi

{\small
\bibliographystyle{ieee}
\bibliography{egbib}
}

\end{document}